\documentclass[journal,twocolumn]{IEEEtran}

\usepackage{booktabs}
\usepackage{xcolor}

\usepackage{comment}

\ifCLASSINFOpdf
   \usepackage[pdftex]{graphicx}
\else
   \usepackage[dvips]{graphicx}
\fi

\usepackage{amsmath}

\newcommand{\networkname}{\textit{RelapsePredNet\:}}
\newcommand{\handcraftedmodel}{\textit{ClusterRFModel\:}}

\begin{document}
\title{Psychotic Relapse Prediction In Schizophrenia Patients Using A Mobile Sensing-based Supervised Deep Learning Model}

\author{Bishal Lamichhane, Joanne Zhou,
        Akane Sano
\thanks{B. Lamichhane and A. Sano are with the Department
of Electrical and Computer Engineering, Rice University, Houston, USA. J. Zhou is with Department of Statistics, Stanford University, California, USA }}

\newcommand\AS[1]{\textcolor{red}{AS:#1}}
\def\BL{\textcolor{blue}}
\def\BLcomment{\textcolor{green}}

\maketitle

\begin{abstract}

Mobile sensing-based modeling of behavioral changes could predict an oncoming psychotic relapse in schizophrenia patients for timely interventions. Deep learning models could complement existing non-deep learning models for relapse prediction by modeling latent behavioral features relevant to the prediction. However, given the inter-individual behavioral differences, model personalization might be required for a predictive model.
In this work, we propose RelapsePredNet, a Long Short-Term Memory (LSTM) neural network-based model for relapse prediction. The model is personalized for a particular patient by training using data from patients most similar to the given patient. Several demographics and baseline mental health scores were considered as personalization metrics to define patient similarity. We investigated the effect of personalization on training dataset characteristics, learned embeddings, and relapse prediction performance. We compared RelapsePredNet with a deep learning-based anomaly detection model for relapse prediction. Further, we investigated if RelapsePredNet could complement ClusterRFModel (a random forest model leveraging clustering and template features proposed in prior work) in a fusion model, by identifying latent behavioral features relevant for relapse prediction. The CrossCheck dataset consisting of continuous mobile sensing data obtained from 63 schizophrenia patients, each monitored for up to a year, was used for our evaluations. The proposed RelapsePredNet outperformed the deep learning-based anomaly detection model for relapse prediction. The F2 score for prediction were 0.21 and 0.52 in the full test set and the Relapse Test Set (consisting of data from patients who have had relapse only), respectively. These corresponded to a 29.4\% and 38.8\% improvement compared to the existing deep learning-based model for relapse prediction. Patients’ SFS score was the best personalization metric to define patient similarity. RelapsePredNet complemented the ClusterRFModel as it improved the F2 score by 26.1\% with a fusion model, resulting in an F2 score of 0.30 in the full test set. This F2 score is the best relapse prediction performance yet in the CrossCheck dataset.

\end{abstract}


\IEEEpeerreviewmaketitle

\section{Introduction}



About 20 million people worldwide are estimated to be suffering from schizophrenia~\cite{James2018}, a chronic mental disorder characterized by psychosis i.e., a symptomatic phase with an abnormal assessment of reality and incidences of hallucinations. Schizophrenia can have heterogeneous presentations across individuals with varying symptoms such as disorganized speech, social withdrawal, paranoia, reduced emotions, etc. Though the prevalence of schizophrenia is lower compared to other mental health disorders such as depressive disorders and bipolar disorders~\cite{kiadaliri2018global}, the effects of schizophrenia can be highly debilitating. It is one of the top 15 leading causes of disability worldwide ~\cite{naghavi2017global} and there is a high risk of self-harm and violence associated with schizophrenia. Relative risks of suicide, for example, have been reported to be one of the highest in schizophrenia patients ~\cite{moitra2021estimating}. Proper management of schizophrenia with therapeutics is required to improve a patient's mental health and facilitate social rehabilitation. Despite being under active treatment, an individual might still show resurgent or exacerbating psychotic symptoms, referred to as an incidence of psychotic relapse in schizophrenia.  

A clinician keeps track of a schizophrenia patient's symptoms using questionnaires such as the Brief Psychiatric Rating Scale (BPRS)~\cite{overall1988brief} during the patient's routine clinical visit. However, clinical visits happen only every few months. A patient might still experience psychotic relapses between the visits. Such relapses create a significant burden on the patient's day-to-day life and might even pose a danger to the life of the patient and their caregivers/family. An automated method of symptoms assessment by modeling a patient's behavior could be useful to predict an oncoming relapse since behavioral changes have been associated with an oncoming relapse~\cite{bouhlel2012prodromal}. A relapse prediction model could inform clinicians/families of necessary interventions and prevent undesirable outcomes from relapse.

Mobile sensing using smartphones has been demonstrated to capture an individual's behavioral patterns~\cite{wahle2016mobile,place2017behavioral,wang2018}. Different sensor modalities readily available in smartphones such as accelerometer, ambient light level, ambient sound level, GPS sensors, etc. can be employed to model behaviors, either explicitly or implicitly ~\cite{wang2018sensing}. For example, the patterns of distance traveled derived from the GPS sensors could inform if an individual traveled for work, went shopping, has been at home, etc. A factor to be considered in behavioral modeling using mobile sensing data is the presence of inter-individual and intra-individual differences in behaviors. Variations in daily behaviors within and across individuals could be due to several person-specific and situation-specific factors such as personality, nature of work, health status, weather, etc. In the context of behavioral modeling for relapse prediction, therefore, a personalized relapse prediction might be required which better represents behavioral patterns associated with normal and relapse behaviors of a given patient. Typical behavioral patterns of a patient could however be shared by a group of other patients. For example, it has been observed in previous works that demographic factors such as age and baseline mental health scores are relevant factors associated with relapse-related behaviors in schizophrenia patients~\cite{alphs2016factors}. 

\section{Related Work}

Some previous works have investigated mobile sensing solutions for relapse prediction in schizophrenia. The authors in ~\cite{Barnett2018} used mobile sensing data obtained from schizophrenia patients for up to 3 months period and identified that mobility patterns and social behaviors could indicate anomalies preceding a relapse. In a follow-up work~\cite{henson2021anomaly}, the detected anomalies were used to predict if a patient has had a relapse. A patient-level classification (classify if a given patient has had a relapse or not), while informative, does not provide a directly actionable prediction such as the prediction of an oncoming relapse.

The authors in \cite{Wang2020} proposed a machine learning-based relapse prediction model using mobile sensing data. Different features such as activity, distance, number of communications, etc. were used in a support vector machine (SVM)-based relapse prediction model. The authors reported an F2 score of 0.28 (F1 score of 0.27). However, the work used a random k-fold cross-validation model for evaluations. Thus, the assessment did not address if relapse prediction based on mobile sensing would be predictive for patients unseen by the model or be predictive in a sequential prediction setting (only data from the past can be used to train/re-train models) encountered in real-life deployments. The efficacy of mobile sensing-based relapse prediction models on unseen patients was evaluated by the authors in ~\cite{lamichhane2020patient,zhou2021routine}. A random forest model for relapse prediction using template-based and clustering-based features derived from different mobile sensing data was proposed. A non-deep learning model such as random forest is constrained by the behavioral representation provided by the input feature as new features cannot be learned automatically towards the relapse prediction task. 


Recently, deep learning models have been successful in many prediction problems~\cite{lecun2015deep}, some of them in the mobile sensing context also~\cite{servia2017mobile,yao2017deepsense,bahador2021deep}. Deep learning models could also be potentially employed for relapse prediction. These models could automatically learn the relevant features using input mobile sensing data towards appropriate behavioral representations required in relapse prediction. Deep learning models for relapse prediction have been explored in previous work. An encoder-decoder neural network model~\cite{sutskever2014sequence,malhotra2016lstm} in an anomaly detection setting was proposed by the authors in ~\cite{adler2020predicting} for relapse prediction. The encoder-decoder model learned abstract representations (embeddings) of the input hourly averages of mobile sensing data obtained from the non-relapse (healthy) days only, during training. Then, an anomaly detection model based on the distances between embeddings was learned in a training set comprising data from non-relapse and relapse phases. The trained model detected more anomalies near the relapse period in the test set. The encoder-decoder model used by the authors is an unsupervised representation learning technique. A supervised deep learning model using mobile sensing data for relapse prediction has not been investigated in previous works though supervised deep learning models have shown promising results in many other mobile sensing-based health applications~\cite{jaques2017predicting,yao2017deepsense,rozet2019using,faruqui2019development,sathyanarayana2016sleep}.

\section{Goal of this Study}


In this work, we aim to investigate a supervised deep learning model for relapse prediction using mobile sensing data. A deep learning model could learn new feature representations to complement existing approaches to relapse prediction using non-deep learning models. We evaluate the relapse prediction model in a patient-independent setting, with sequential prediction. This provides a better representation of how a relapse prediction model would be deployed. As there might be large inter-individual differences in behavioral patterns, we explore model personalization by training a model for a test patient using data from other patients most similar to the given test patient. Accordingly, we aim to evaluate different demographic and mental health questionnaire scores to define the patient similarity. Finally, we aim to compare the supervised deep learning model for relapse prediction with a deep learning-based anomaly detection model for relapse prediction which uses unsupervised representation learning.

\section{Methods}

\subsection{Dataset}


We used the Crosscheck dataset~\cite{Wang2016, BenZeev2017} for the development and evaluation of the relapse prediction model. The crosscheck dataset was obtained from a study on schizophrenia patients conducted at the Zucker Hillside Hospital, New York City (Clinical Trial Registration Number: NCT01952041). The study was approved by Dartmouth College's ethical review committee (\#24356) and the institutional review board at North Shore-Long Island Jewish Health System (\#14-100B). Patients who enrolled in the study provided written informed consent. Further details about the study such as inclusion criteria have been described in~\cite{BenZeev2017}. A total of 63 outpatients completed the study, where continuous mobile sensing data was obtained from each of the participants for over a year. Several pre-study surveys providing a comprehensive assessment of the baseline symptoms and functioning of the patients are also available in the dataset. The demographics of the patients were as follows: 27 male/36 female and an average age of 37.2 $\pm$ 13.7 years. Various mobile sensing data obtained from the smartphone sensors, either directly (e.g., accelerometer) or derived (e.g., conversation duration), are available in the Crosscheck dataset. The dataset also contains information about the dates when the patients experienced a relapse. A patient was registered to have relapsed based on criteria such as psychiatric hospitalization, the need for increased clinical care, increased BPRS scores, etc. Of the 63 patients who completed the study, 20 patients had a relapse with a total of 27 instances of relapse registered. Some patients had multiple relapses during the monitoring period.

\subsection{Features}

We used the mobile sensing data in the CrossCheck dataset for behavioral modeling toward relapse prediction. In particular, we used the data from the following six mobile sensing modalities in this work:

\begin{enumerate}
    \item Light exposure (Light) - average light exposure magnitude
    \item Volume (Vol) - average audio amplitude
    \item Conversation (Conv) - duration of detected conversations
    \item Distance (Dist) - total distance traveled
    \item Accelerometer (Acc) - average accelerometer magnitude 
    \item Screensum (Screensum) - total screen usage
\end{enumerate}

These mobile sensing data were measured/inferred with varying sampling frequencies. Hourly averages for each of the six modalities were computed to obtain a six-dimensional feature vector representing each hour of the measurement period. The entire day is thus represented by a 144-dimensional feature vector (24 hours x 6 feature vectors/hour).

\subsection{Relapse Prediction Model}

\begin{figure}
    \centering
    \includegraphics[width=0.3\textwidth]{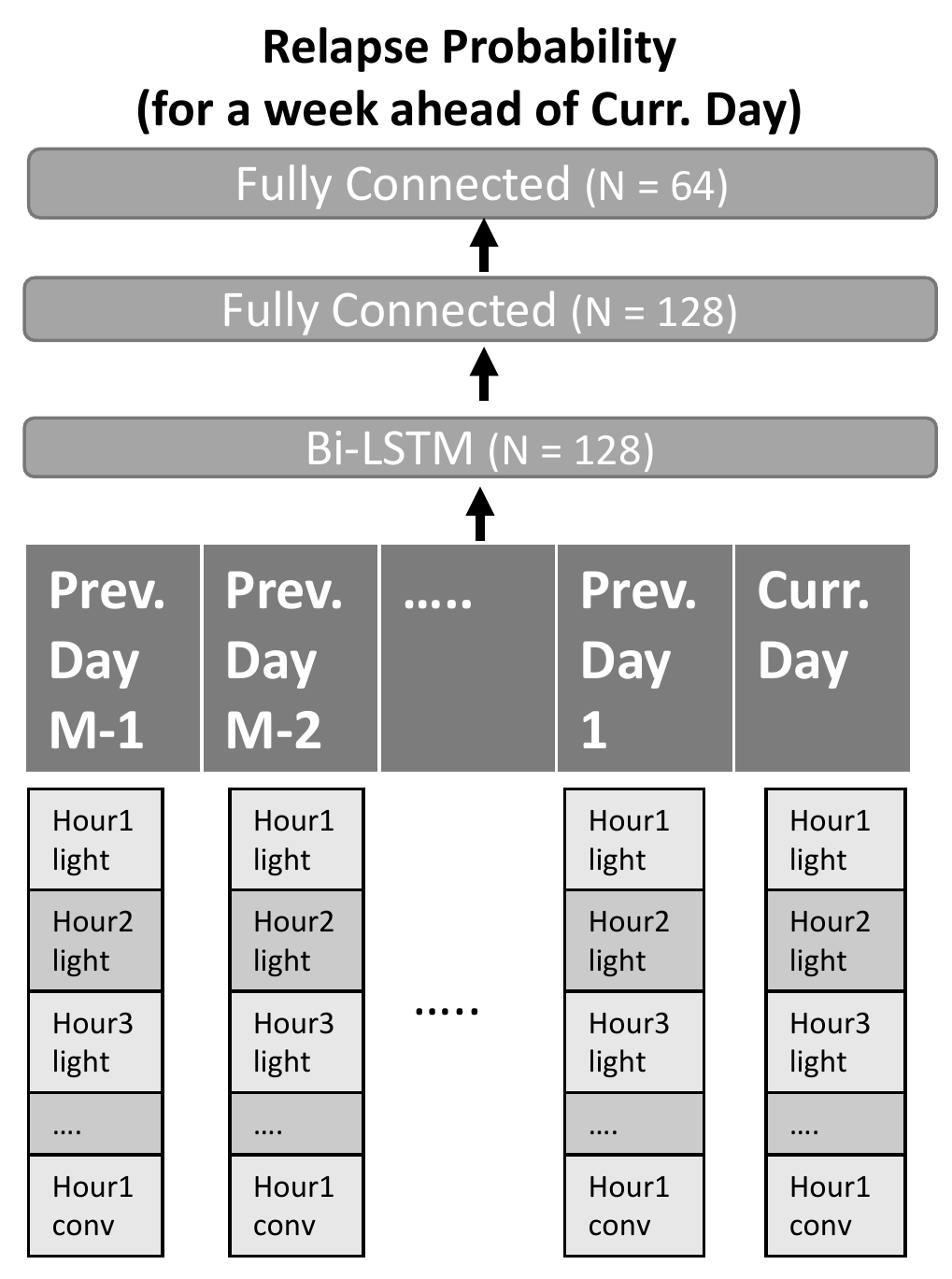}
    \caption{
    Relapse prediction model, \networkname: A 128 unit bi-directional LSTM model (Bi-LSTM) is used to learn patterns differentiating relapse from non-relapse using the temporal progression of hourly sensor data across days, with data from the past M days used in this prediction. The model input is 144-dimensional mobile sensing data consisting of hourly data from 6 mobile sensing modalities.}
    \label{fig:mh_lstm_model}
\end{figure}

With 144-dimensional hourly mobile sensing data representing a day, we developed a relapse prediction model based on the behavioral representation provided by this mobile sensing data. We used an LSTM-based deep learning model for relapse prediction. We refer to the model as \networkname. The architecture of \networkname is shown in Figure~\ref{fig:mh_lstm_model}. The bi-directional LSTM layer learns features differentiating relapse from non-relapse using the temporal progression of hourly sensor data across days. The activations from the LSTM layer are then passed onto two successive fully connected layers to obtain a relapse prediction probability.  A 128-unit bidirectional LSTM model and fully connected layers of sizes 128 and 64 were used for \networkname. We used batch normalization between the LSTM layer and the first fully connected layer as well as between the first and second fully connected layers. Similarly, a dropout on connections from the LSTM layer to the first fully connected layer was used, with a dropout rate of 0.2, for model regularization. This network architecture was chosen to have a smaller network (given the limited dataset available to train) while still having a good prediction performance, as assessed by cross-validation.  

We used a sequential prediction model within a leave-one-patient-out cross-validation setting to evaluate relapse prediction performance. The patient for which the relapse prediction performance is to be evaluated was reserved as the test set. \networkname was trained on the data from the remaining patients in the training set. A previous study has shown that the behavioral aberrations related to an oncoming relapse could be present up to a month before a relapse happens~\cite{Wang2020}. Therefore, we used 4 weeks of data to predict an oncoming relapse in the next week with a trained model. The feature extraction/prediction window was moved sequentially by one week (sliding window with a step size of one week). Thus, \networkname produced a prediction for every week of monitoring. The patient-independent evaluation setting with sequential prediction better represents how a model would be deployed in clinical practice. Earlier works such as \cite{Wang2020, adler2020predicting} evaluated relapse prediction performance with random k-fold cross-validation. Such evaluation setup implies that not only the patient on which the relapse prediction model is evaluated has been represented in the training set but even the data from the future (later months of monitoring data) is likely used for predictions for the current period.  

As prescribed in earlier works~\cite{adler2020predicting,zhou2021routine}, we labeled each week within a month of relapse as a positive label for relapse. A prediction of an oncoming relapse within a month of an actual relapse could be valuable for interventions and likely to be assignable to behavioral changes in patients. We trained \networkname using binary cross entropy (BCE) loss on the relapse/non-relapse binary label with the ADAM optimizer. The learning rate was set to 0.00001 and the batch size was set to 32. 

\subsection{Personalization}




The patient-independent evaluation with leave-one-patient-out cross-validation allows us to assess the efficacy of relapse prediction on an unseen patient. However, there could be large inter-individual differences in behaviors. A generalized relapse prediction model (trained on a population and not specific to a particular patient or patient-groups) might thus have limited predictive value. Accordingly, we investigated the use of a personalized relapse prediction model. 


\subsubsection{\networkname With Personalization}
\label{sec:no_personalization_baseline}

We developed a personalized relapse prediction model by training the model for a particular patient (test patient) using the data from other patients that are most similar to the given test patient. The similarity between patients was defined based on either demographic characteristics of the patients or their baseline mental health scores (assessed using clinical questionnaires in the pre-study surveys). The patient characteristic used to define patient similarity and subsequently drive the personalization process is referred to as the \textit{personalization metric}. In this work, we evaluated the following patient characteristics as possible \textit{personalization metric}:



\begin{itemize}
    \item Patient's Age
    \item Baseline BPRS~\cite{overall1988brief} scores 
    \item Baseline Social Functioning Scale (SFS)~\cite{birchwood1990social} scores 
    \item Baseline Calgary Depression Scale for Schizophrenia (CDSS)~\cite{addington1990depression} scores
    \item Baseline Green et al., Paranoid Thoughts Scale (GPTS)~\cite{green2008measuring} scores 
    \item Combined metric (each of the previous metrics scaled to 0-1 with min-max scaling and averaged)
\end{itemize}

Let $<x_{i,j},y_{i,j}>$ represent the training data and label pairs for relapse prediction where $i,j$ are the index of the data sample and patient respectively. If $<x_{i,T},y_{i,T}>$ are the data from the test patient $T$, we identify the personalization subset $<x_{i,\{K_1,..,K_N\}},y_{i,\{K_1,..K_N\}}>$ in the training set consisting of the data from patients ${K_1,...,K_N}$. These patients ${K_1,...,K_N}$ are most similar to the test patient $T$, based on the similarity defined using a \textit{personalization metric}. For example, if age is used as the \textit{personalization metric} then patients ${K_1,...,K_N}$ have the smallest absolute age difference from patient $T$'s age. Since the instances of relapse are relatively rare, all the instances of relapse in the training dataset are included in the personalization subset. If $N_{rel}$ is the number of relapse instances in the personalization subset then equal number of non-relapse instances $N_{nonrel} = N_{rel}$ are included to have a balanced dataset. The relapse prediction model for patient $T$ is then trained using hence constructed balanced personalization subset.

\subsubsection{\networkname Without Personalization}
\label{sec:no_personalization_baseline}

To assess the need for personalization, we also evaluated the relapse prediction from \networkname when no personalization is used. For this, the \networkname is trained with the entire training set. As the training set thus constructed is highly imbalanced, we evaluated models trained with both the BCE loss and F2 loss. The latter directly optimizes the F2 score (Section~\ref{sec:evaluation_metric}) and could be suitable for learning in an imbalanced dataset.

\subsubsection{\networkname With Randomly Sampled Training Set}
\label{sec:randomsampling_baseline}

Besides the non-personalized relapse prediction model, we also evaluated models trained with randomly sampled training sets. The sampled training set is balanced, similar to personalized model training. The model training based on random sampling is evaluated for ten independent runs and the average prediction performance is reported.

\subsubsection{Unimodal \networkname}

For the best \networkname model (using the identified best \textit{personalization metric}), we investigated the prediction power of unimodal \networkname  i.e., \networkname trained on a single sensor modality. Unimodal model evaluations would give insights on the most relevant modality for relapse prediction and guide future investigations on modality combination strategies for improved relapse prediction.

\subsection{Fusion Model}

\begin{figure*}
    \centering
    \includegraphics[width=1.6\columnwidth]{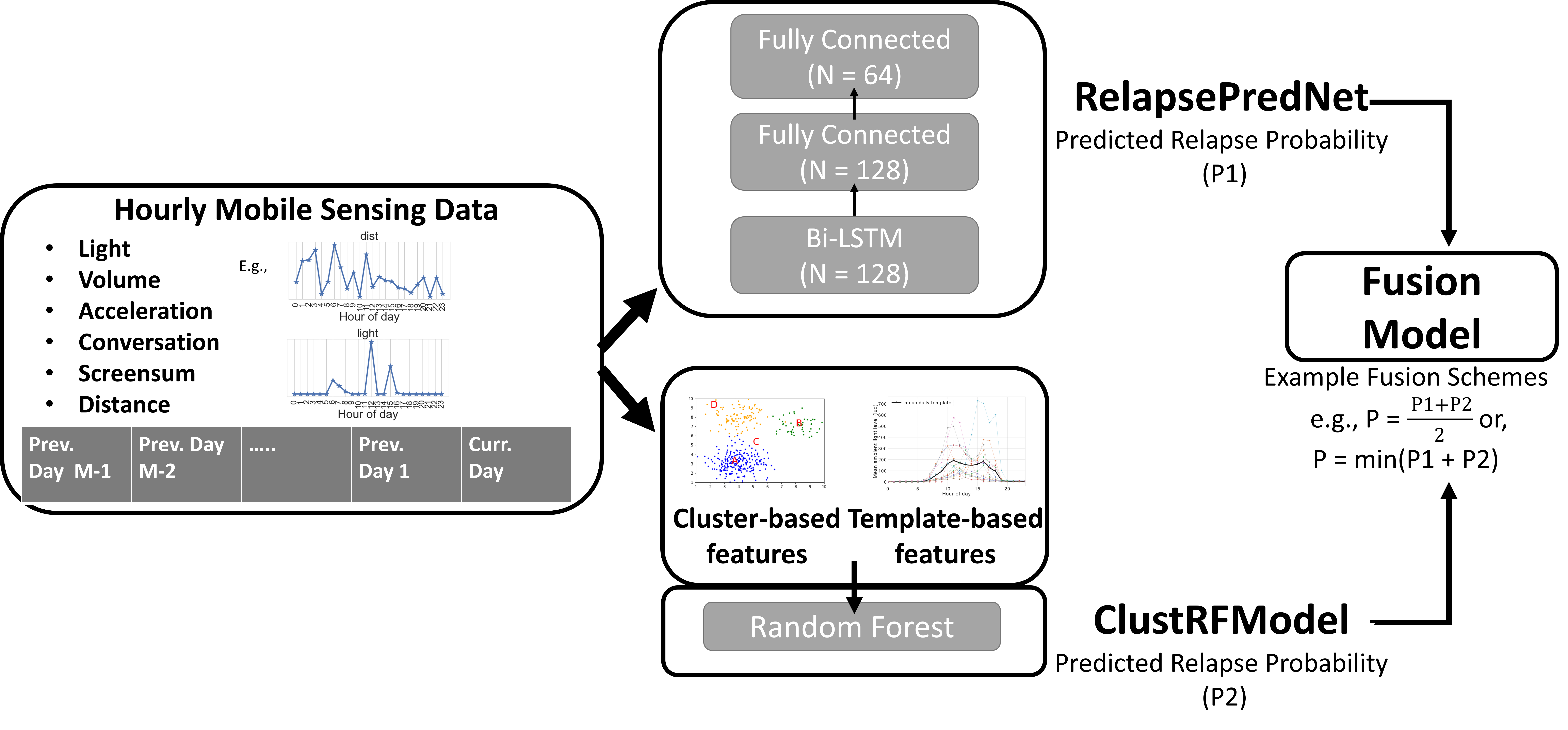}
    \caption{Fusion model where the relapse prediction output from \networkname is combined with the output from the \handcraftedmodel, a random forest-based model leveraging clustering and template-based features proposed in a prior work~\cite{zhou2021routine}. While both \networkname and \handcraftedmodel use hourly mobile sensing data as their input, the \networkname relies on supervised deep learning technique to \textit{identify} features automatically. The learned features could be complementary to the template-based and clustering-based features for the relapse prediction task.}
    
    \label{fig:fusion_model}
\end{figure*}

Deep learning models for relapse prediction could be complementary to non-deep learning models as new latent features relevant to the prediction task could be learned automatically based on the input mobile sensing data (hourly mobile sensing data in context of our work). To evaluate this, we assessed the prediction performance of a fusion model. The fusion model combines \networkname with the \handcraftedmodel, a random forest-based model proposed by the authors in~\cite{zhou2021routine} which leverages template and clustering-based mobile sensing features. This fusion model is demonstrated in Figure~\ref{fig:fusion_model}.

We explored late fusion approaches where the output from the models is combined to obtain a final assessment. We evaluated different late fusion schemes based on how the relapse probability from the two models is fused. Particularly, we evaluated the \textit{mean, min, and max} fusion models where we considered the average, minimum, or the maximum of the two output probabilities respectively.    



\subsection{Baseline Models}

In this work, we evaluated two baseline models for relapse prediction. The first baseline model is the encoder-decoder model of \cite{adler2020predicting} employed in an anomaly detection setting. The second baseline model is a Random Forest (RF) model.

For the encoder-decoder model, we used the same fully connected deep neural network as proposed in the work of \cite{adler2020predicting}, along with the proposed anomaly detection setting. The encoder-decoder model is trained to reconstruct the input hourly mobile sensing data from a day. The encoder output from the trained model serves as an embedding of the mobile sensing data. The anomaly detection is then based on the Mahalanobis distance of an embedding from the distribution of embeddings from non-relapse phases. The optimal distance threshold differentiating embeddings from relapse and non-relapse phases is learned from the training set. Thus, though the representation learning of embeddings is unsupervised, some supervision is involved in learning the optimal threshold separating relapse instances (anomalies) from the non-relapse instances (normal data). 

For the RF model, we obtained mean (across time-axis) mobile sensing data within a feature extraction window (Figure~\ref{fig:mh_lstm_model}) and provided these as input features to the RF model. We used an RF model with 11 decision trees in the ensemble; a higher number of trees were not translating to improved prediction performance. 

We trained the baseline models with personalization similar to how \networkname is trained.

\subsection{Evaluation Metric}
\label{sec:evaluation_metric}

We evaluated the relapse prediction performance using $F2$ score which is defined as: 
\begin{equation*}
    F2 = \frac{5 * \text{precision} * \text{recall}}{4 * \text{precision} + \text{recall}}
\end{equation*}

An $F2$ score prioritizes recall over precision, therefore assigning more importance to detecting an oncoming relapse compared to the costs associated with a generated false alarm (false positive). F2 score has been used in previous work for relapse prediction evaluation~\cite{lamichhane2020patient,zhou2021routine}. For all evaluations, we report the average F2 score obtained from ten different runs (with different random initialization), unless mentioned otherwise. 


\subsection{Visualization of Sensor Data in Relation with Personalization Metric}
\label{sec:visualization}


To understand the role of personalization in relapse prediction models, we compared the intra-class distances of the hourly mobile sensing data (input data) with their inter-class distances. The intra-class distances are the distance between the data points within the non-relapse class (NonRelapse-NonRelapse distances). Similarly, the inter-class distances are the distance between the data points in the non-relapse class and data points in the relapse class (NonRelapse-Relapse distances). A larger difference between the intra-class distances and the inter-class distance indicates a likely higher separability. We computed the NonRelapse-NonRelapse distances and the NonRelapse-Relapse distances in two cases: with and without personalization. With personalization, we sub-sampled the training set to retain data from patients who are closer (based on \textit{personalization metric} distance) to the test patient. This is compared with the no personalization case where the training set is sub-sampled randomly. We visualized the NonRelapse-NonRelapse and NonRelapse-Relapse distances with an example subject in the test set with a high F2 score (F2 = 0.83). This gives a better contrast in differences between the two distances, i.e., the intra-class and inter-class distances, if any. We also analyzed the embeddings (output from the last hidden layer of the neural network model) from a trained \networkname for the two different training settings, i.e., training with a personalized subsampling versus training with a random subsampling.

The intra-class and inter-class distances were computed as follows. 
For each data point in the training set obtained according to the two subsampling approaches - personalized and random - we computed the average mobile sensing feature vector in a given feature extraction window (averaging across time-axis). The average feature vector representing a feature extraction window is thus a 144-dimensional vector. Subsequently, we obtained Euclidean distance between the normalized feature vectors (normalized with a min-max scaler) to visualize the distribution of distances in the NonRelapse-NonRelapse and NonRelapse-Relapse categories. We also obtained the activations - or embeddings - from the 64-dimensional hidden layer of \networkname for the data points in the test set. We visualized the embeddings using their 2-dimensional t-SNE (T-distributed Stochastic Neighbor Embedding~\cite{van2008visualizing}) projections. The separability of the projected points was quantified using the separability index~\cite{thornton1998separability}, a higher separability index indicating that points are more likely separable, and silhouette coefficient where a higher coefficient indicates a better distinguishable cluster.

\section{Results}

\subsection{\networkname Without Personalization}

We evaluated the relapse prediction from \networkname when no personalization was used or when the training set was randomly sampled to have a dataset comparable to the personalized subset.  The results obtained with these different model training strategies are presented in Table~\ref{tab:personalization_nopersonalization}. The relapse prediction model trained on the entire training dataset has no predictive performance (F2 score of 0). A loss function sensitive to the data imbalance was also not able to improve predictive performance. However, random subsampling (undersampling in the imbalanced classification terminologies) provided some predictive performance in the model, though the obtained F2 score was still low. 

\begin{table}[!htb]
    \centering
        \caption{Relapse prediction from \networkname without personalization, i.e., when the entire training dataset was used for model training. As the training set was imbalanced, we evaluated model trained with both the BCE loss (as used in \networkname under personalization) and the F2 loss (suitable for learning in an imbalanced dataset scenario). }
    \begin{tabular}{c|c}
    \toprule
    \textbf{Model} & \textbf{F2} \\
    \midrule
    \shortstack{\networkname Without Personalization \\- BCE loss, Full Training Set} &  0.000 \\
    \hline
    \shortstack{\networkname without Personalization \\- F2 loss, Full Training Set} & 0.013 \\
    \hline
    \shortstack{\networkname without Personalization \\- BCE loss, randomly subsampled training set} & 0.104  \\
    \bottomrule
    
    \end{tabular}
    \label{tab:personalization_nopersonalization}
\end{table}

\subsection{Personalized Relapse Prediction: Comparison of Different Personalization Metric}

\begin{figure}
\centering
    \includegraphics[width=0.7\columnwidth]{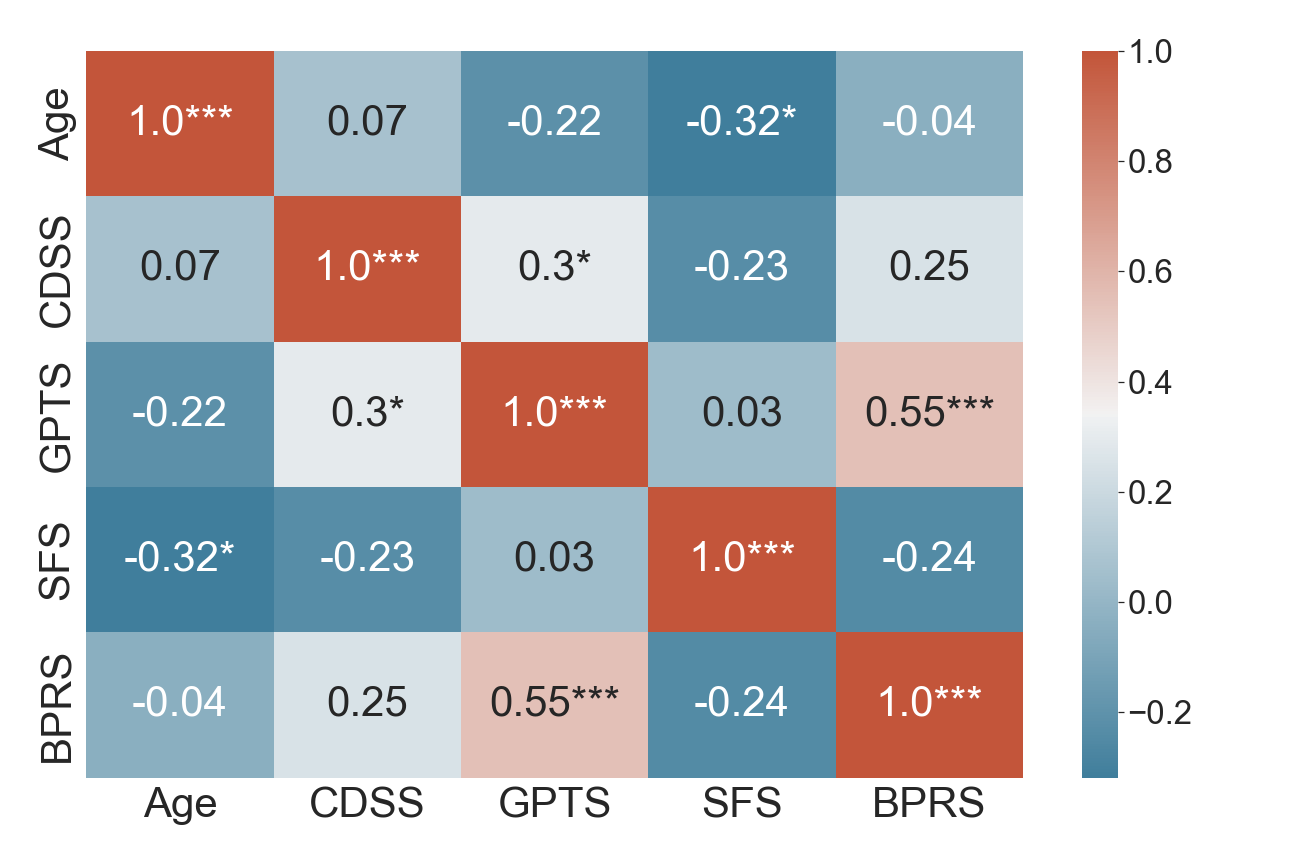}
    \caption{Correlations between different \textit{personalization metrics} for the patients in the CrossCheck dataset. * indicates P<.05, ** P<.01, and *** P<.001. }
    \label{fig:corr_personalization_metric}

\end{figure}

The personalization method used in \networkname requires the identification of patients who are similar to the given test patient. We defined patient similarity with \textit{personalization metrics} i.e., different demographic characteristics and baseline mental health scores of the patients. The correlation between the \textit{personalization metrics} evaluated in this work is shown in Figure~\ref{fig:corr_personalization_metric}. The \textit{personalization metrics} were found to be correlated with each other. Age had a significant negative correlation with SFS score. Similarly, the CDSS score was positively correlated with the GPTS score and there was a significant positive correlation between GPTS and BPRS score. 

\begin{figure}
    \centering
    \includegraphics[width=0.8\columnwidth]{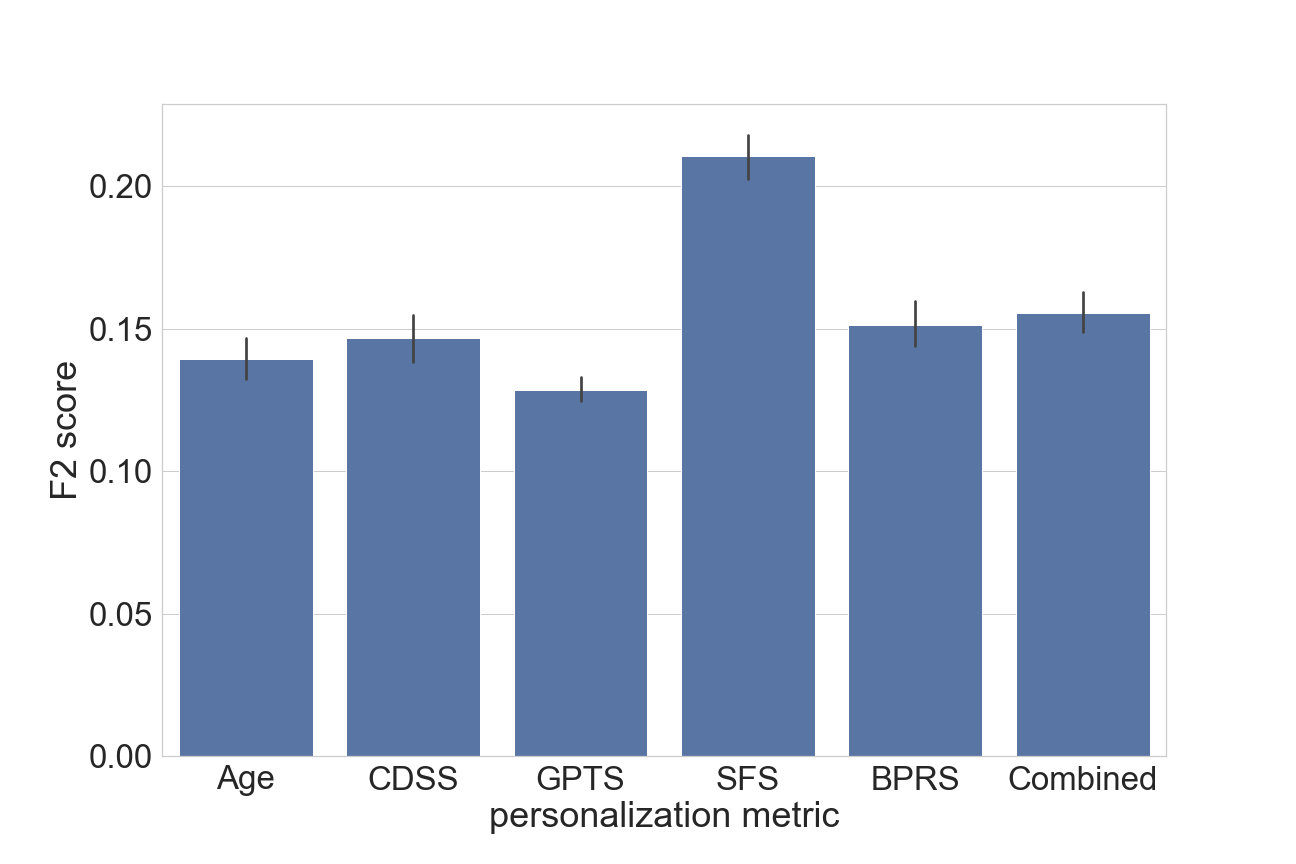}
    \caption{Relapse prediction with personalization based on different personalization metrics. Demographic information such as age and baseline mental health status such as BPRS score at intake are used to identify a personalization subset where the relapse prediction model is trained.}
    \label{fig:diff_personalization_metric}
\end{figure}

We evaluated the relapse prediction performance from \networkname when various \textit{personalization metrics} are used to obtain the personalization subset. The result obtained is shown in Figure~\ref{fig:diff_personalization_metric}. The best prediction performance was obtained when patient similarity is defined based on their SFS score at intake. The F2 score obtained with SFS as the personalization metric was 0.21. The F2 score under any of the personalization metrics was still better than the F2 score obtained with random subsampling (undersampling).  

Of the 63 patients in the Crosscheck dataset, 20 patients have had relapse(s). We calculated F2 scores for these relapse patients (since F2 scores are undefined for non-relapse patients where no positive class labels are present) using the \networkname trained with SFS-based personalization. The average F2 score per patient, when only patients who have had relapse were considered, was 0.37. The F2 score was lower when the entire patient cohort was considered because of the false alarms generated in patients who had no relapse.

\begin{figure}
    \centering
    \includegraphics[width=0.7\columnwidth]{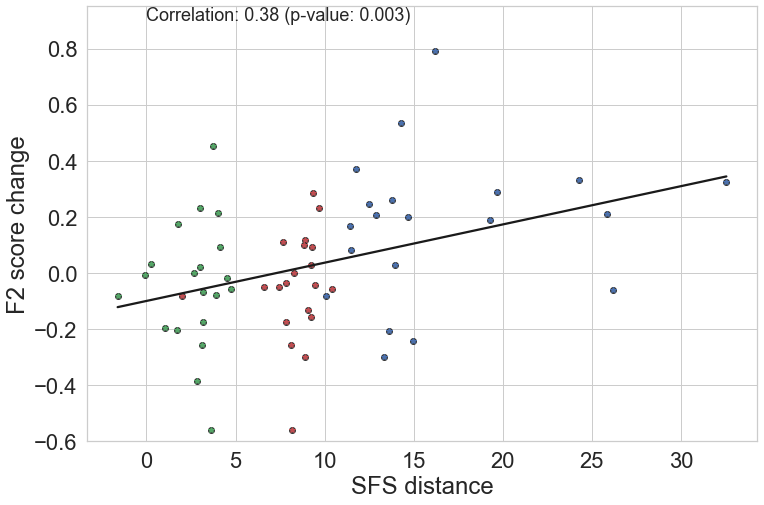}
    \caption{The effect of SFS-based personalization on relapse prediction when the personalization subset is at different distances from the test set. We consider the performance obtained under random subsampling as a baseline. Then the performance obtained when the personalization subset is closest (finding the nearest patients in terms of SFS distance), at the first quartile (in terms of SFS distance), and at a median distance is obtained. The changes in the F2 score, compared with the baseline, for different SFS distance changes, are shown. The SFS distance change is highest when the closest subset is considered, compared to the baseline. The F2 score improvement associated with it is also among the highest. Overall, a significant positive correlation between SFS distance and F2 score improvement is obtained.}
    
    \label{fig:sfsDisteffect}
\end{figure}

\begin{figure}
    \centering
    \includegraphics[width=0.7\columnwidth]{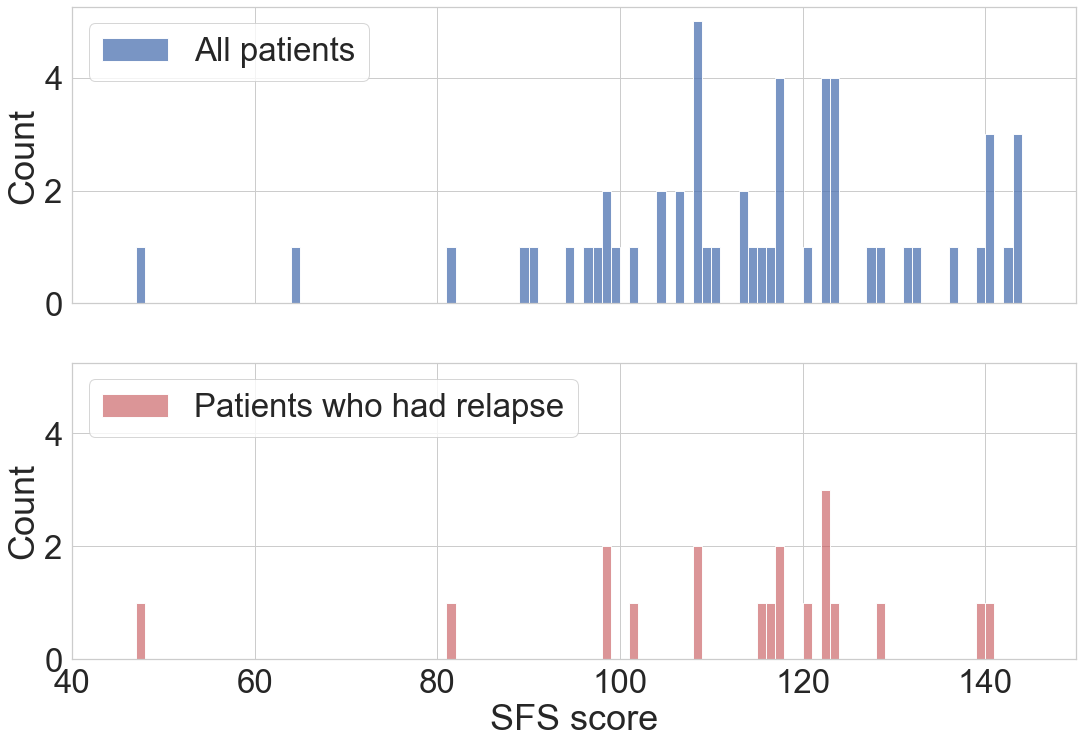}
    \caption{Histograms of SFS scores for patient in the Crosscheck dataset, The top plot shows the histogram for all patients in the dataset. There are patients who have same SFS scores as other other patients while there are also patients who have unique SFS scores. When a personalization subset for a patient in the test set is constructed, the data from other patients who are closest to the given patient are used as the tranining set. The bottom plot shows the histogram for patients who have had relapse(s).}
    \label{fig:sfs_histogram}
\end{figure}

To confirm the effect of personalization on relapse prediction performance, for the 20 patients who had a relapse, we analyzed the relation of relapse prediction performance with the average \textit{personalization metric} distance (absolute difference in SFS scores) between the participants in the training set and test set. We obtained a personalization subset composed of patients in the training set that are closest in terms of \textit{personalization metric} distance to the patient in the test set. We then considered the relapse prediction performance obtained when the personalization subset was composed of patients at different \textit{personalization metric} distances from the patient in the test set. The performance obtained under random subsampling (where we randomly sampled matched patients and the average performance from ten runs) was considered as the baseline. 
Let $F2$ and $dist$ be the F2 score and average SFS distance between training set patients and the test set patient. Similarly, let $F2_{rand}$ and $dist_{rand}$ be the average F2 score and the average \textit{personalization metric} distance between the patients in the training set and test set in the baseline case, respectively.
We computed the change in relapse prediction performance, compared to the baseline, when the personalization subsets were at different \textit{personalization metric} distances from the test set. Particularly, we obtained relapse prediction performance when the personalization subset was composed of patients that are \textit{closest}, (composed of patients with the least SFS distance to the test patient), \textit{at the first-quartile} (composed of patients whose distances to the test patient start after the first quartile of all training set patients to test set patient distances), and \textit{at a median} distance. Specifically, we analyzed the relation of $dist_{rand} - dist$ and $F2 - F2_{rand}$. This relation would characterize how much improvement in F2 score (if any) is brought by increasing closeness of the training set patients and the test set patient, for a given patient. The correlation of SFS distance changes with F2 score changes was 0.38 ($p$ =  0.003), as shown in Figure~\ref{fig:sfsDisteffect}. The positive correlation indicates that the F2 score improves as the personalization subset is closer to the patient in the test set (larger distance changes). The distribution of SFS scores for the patients in the Crosscheck dataset is shown in Figure~\ref{fig:sfs_histogram}. For a selected patient, the personalization subset could be constructed from other patients who have the same or most similar SFS scores as the given patient, as we proposed for \networkname with personalization. An improved relapse prediction when the selected personalization subset was closer to the test patient indicates that personalization helps improve prediction.

\subsection{Comparison of Different Modalities for Relapse Prediction: Unimodal \networkname}

We used the data from six mobile sensing modalities as input in \networkname to have different behavioral representations captured by these modalities. For example, conversation and volume represent social interaction and sociability (proclivity to be around other people rather than be isolated) in general. Similarly, distance and acceleration represent mobility and active/non-active lifestyles. Given that the best performing \networkname was obtained with SFS-based personalization, we evaluated models trained on a single modality using SFS-based personalization. The input feature dimension of these models was 24 (hourly feature for a modality). The result obtained was presented in Table~\ref{tab:pred_single_modality}. The best prediction performance is obtained with the model trained on the conversation modality followed by the model trained on the volume modality. Surprisingly, the model trained on the conversation modality alone performed on par with \networkname trained on all six modalities. 

\begin{table}[!htb]
        \caption{Relapse prediction using features from a single modality. Personalization subsets are still created using SFS-based patient similarity for training with the personalization subset only.}
    \centering
    
	\scalebox{1.0}{
	
    \begin{tabular}{c|c}
        \toprule
        \textbf{Modality} & \textbf{F2}  \\
        \midrule
         Light & 0.16  \\
         \hline
         Distance & 0.17  \\
         \hline
         Conversation & 0.21  \\
         \hline
         Volume & 0.17  \\
         \hline
         Acc & 0.16  \\
         \hline
         Screen Usage & 0.16 \\
         \bottomrule
    \end{tabular}
    
    }
    
    \label{tab:pred_single_modality}
\end{table}

\subsection{Visualization of Mobile Sensing Data and LSTM Embeddings in Relation with Personalization Metric}

\begin{figure}
    \centering
    \includegraphics[width=\columnwidth]{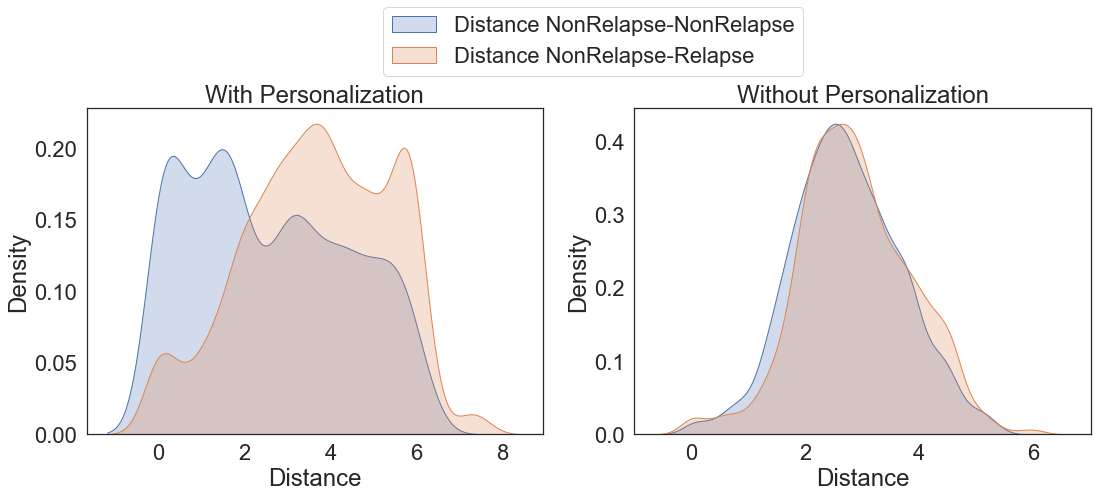}
    \caption{ Distance between non-relapse data points (intra-class distances) and the distance between relapse and non-relapse data points (inter-class distances) in the training set under two different subsampling approaches. The relapse data points have higher distances from the selected non-relapse data points when personalization is used (left) compared to when personalization is not used (right). }
    \label{fig:sensordist_nonrel_rel_distribution}
\end{figure}

To understand why personalization could be helpful for relapse prediction models, we obtained the distribution of NonRelapse-NonRelapse and NonRelapse-Relapse distances under two subsampling strategies: personalized subsampling (the training set was selected based on the patient similarity of their SFS scores) and random subsampling. The result obtained for an example subject in the test set is shown in Figure~\ref{fig:sensordist_nonrel_rel_distribution}. When personalization was used, the distribution of NonRelapse-NonRelapse and NonRelapse-Relapse distances were further apart, in comparison to when personalization was not used.

\begin{figure}
    \centering
    \includegraphics[width=\columnwidth]{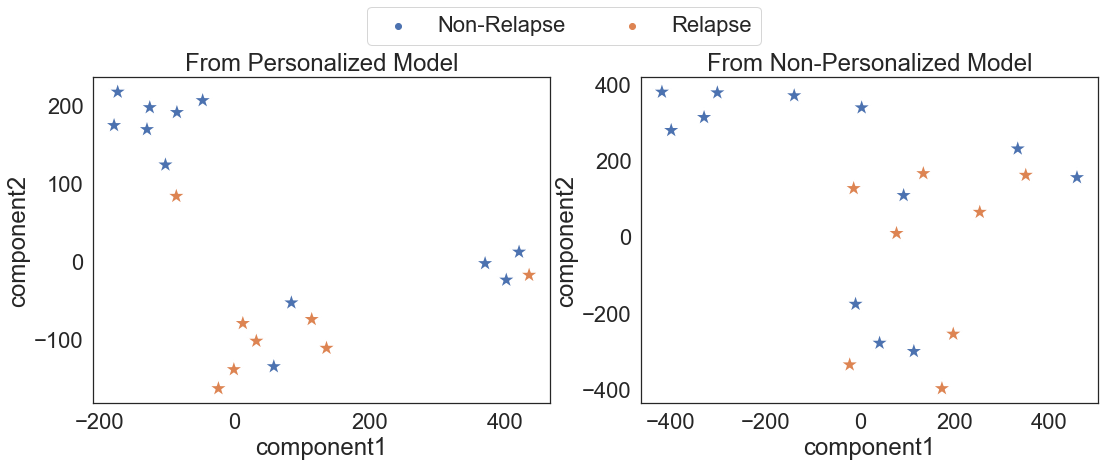}
    \caption{ t-SNE projection obtained for the embeddings from a personalized model (left) and a non-personalized model (right), In the given example, separable clusters with one cluster dominated by relapse data points are obtained from the personalized model.}
    \label{fig:tsne_personal_nonPersonal}
\end{figure}

The trained model, with and without personalization, was used to obtain embeddings for the data points in the test set. The two-dimensional t-SNE projections for these embeddings are shown in Figure~\ref{fig:tsne_personal_nonPersonal}. For the example subject, a personalized model provided more separable clusters (silhouette coefficient: 0.126, separability index: 0.6)) of the relapse and non-relapse data points in the test set in comparison to the model without personalization (silhouette coefficient: 0.118, separability index:0.5). 

\subsection{Comparison of \networkname with baseline model}

We compared the prediction performance of \networkname with those from the baseline models, namely the encoder-decoder model \cite{adler2020predicting} and the Random Forest (RF) model. The results obtained are presented in Table~\ref{tab:model_comparison}. \networkname outperformed both the baseline models.

\begin{table}[!htb]
    \centering
        \caption{Relapse prediction performance from different relapse prediction model architecture, namely encoder-decoder architecture-based anomaly detection model and Random Forest (RF) model}
    \begin{tabular}{c|c}
         \toprule
         \textbf{Model} & \textbf{F2}  \\
         \midrule
         \networkname & 0.21    \\
         \hline         
         Encoder-Decoder based anomaly detection model from \cite{adler2020predicting} & 0.16   \\
         \hline
         RF &  0.15  \\
         \bottomrule
    \end{tabular}
    \label{tab:model_comparison}
\end{table}

\subsection{Relapse Prediction Performance in the Relapse Test Set}

The authors in \cite{adler2020predicting} evaluated relapse prediction in a test set comprising of relapse instances and 20\% of the non-relapse instances from the patients who have had a relapse. We denote this test set as \textit{Relapse Test Set}. We evaluated the relapse prediction performance obtained from \networkname and the encoder-decoder model from \cite{adler2020predicting} in the \textit{Relapse Test Set}. The results obtained are presented in Table~\ref{tab:results_relapsepredset}. The fully supervised deep learning model of \networkname has a higher F2 score even when evaluated in the \textit{Relapse Test Set}.

\begin{table}
\caption{Relapse prediction performance obtained in the \textit{Relapse Test Set} from our proposed supervised prediction model and the encoder-decoder model proposed in \cite{adler2020predicting}}

\centering

\begin{tabular}{c|c}
\toprule
\textbf{Model} & \textbf{F2 score} \\
\midrule
\networkname & 0.52 \\
\hline
Encoder-Decoder based anomaly detection model from \cite{adler2020predicting} & 0.37 \\
\hline

\end{tabular}
\label{tab:results_relapsepredset}
\end{table}

\subsection{Fusion Model For Relapse Prediction}

We evaluated the relapse prediction performance of a fusion model, one that does a late fusion of relapse probability from deep learning-based \networkname and \handcraftedmodel from~\cite{zhou2021routine}, a non-deep learning model based on random forest trained using template and clustering-based features of mobile sensing data. The results obtained for different fusion schemes are presented in Table~\ref{tab:fusion_model_evaluation}. The fusion model taking the minimum of the predicted relapse probability improves the F2 score to 0.30 compared to the F2 score of 0.21 (from \networkname) and 0.23 (from \handcraftedmodel) of the individual models.

\begin{table}[!htb]
    \centering
    \caption{Relapse prediction performance obtained with a fusion model combining \networkname and \handcraftedmodel. Different late fusion schemes, corresponding to how the relapse probability from deep learning-based model and handcrafted features-based model are combined, were evaluated.}

    \begin{tabular}{c|c}
        \toprule
        \textbf{Late fusion scheme} & \textbf{F2 score} \\
        \midrule
         Mean fusion & 0.23  \\
         \hline
         Max fusion & 0.19 \\
         \hline
         Min fusion & \textbf{0.30} \\
         \bottomrule
    \end{tabular}
\label{tab:fusion_model_evaluation}
\end{table}

\section{Discussion}

\subsection{Principal Results}

Relapse prediction from mobile sensing data could facilitate timely interventions and prevent undesirable outcomes due to the psychotic symptoms in schizophrenia patients. In this work, we proposed a personalized relapse prediction model using a supervised deep learning model trained on hourly mobile sensor data. We personalized the prediction model by adapting the model according to the characteristics of the test patient. Several demographic and mental health scores were evaluated as potential \textit{personalization metrics}. The SFS score, measuring the social skills of schizophrenia patients, was the best metric to characterize patients for personalization. Our proposed supervised deep learning model outperformed the existing encoder-decoder network-based anomaly detection model for relapse prediction~\cite{adler2020predicting} and an RF baseline model for relapse prediction. Additionally, the proposed deep learning-based model was found to complement the non-deep learning model for relapse prediction~\cite{zhou2021routine} proposed in previous work.

The relapse prediction problem is class-imbalanced as the incidences of relapse are rare. When \networkname was trained with the entire training set (highly imbalanced), the model had no predictive performance (Table~\ref{tab:personalization_nopersonalization}). A class-imbalance-sensitive loss could not improve the predictive performance. Only with random subsampling of the training set did \networkname provide some predictive power. As the training dataset is smaller compared to the dataset size used to commonly train deep learning models, a balanced dataset possibly allows better training of the model with some generalization.  

Relapse prediction models can be personalized by training the model for a test patient using data from other patients who are most similar to the given test patient. Different demographic and baseline mental health scores could be useful when defining the similarity between patients (Figure~\ref{fig:diff_personalization_metric}). The best relapse prediction performance was obtained when the patient similarity was defined in terms of their SFS score. There was a statistically significant difference in performance across \textit{personalization metrics} (ANOVA F-statistics: 54.77, $p$ $<$ .001). The prediction performance obtained with SFS was significantly higher than that obtained with the next best performing metric BPRS (t-statistics=9.9, $p$ $<$ .001). 
The SFS assesses the sociability of schizophrenia patients. Therefore, two patients with the same/similar SFS score could be expected to have similar social behaviors (e.g., in terms of how often they are around other people, have conversations, are moving, etc.). The model with personalization had a higher F2 score compared to the models without personalization. The use of any demographic or mental health scores was found to have some value towards personalization. This was likely because these demographic characteristics and mental health scores were correlated with each other (Figure~\ref{fig:corr_personalization_metric}). For example, personalization using age and SFS score could likely select a similar subset of patients for the personalization subset since age and SFS score were significantly correlated. 

That the personalization was helpful for relapse prediction was further demonstrated when we obtained improvement in F2 score as the SFS score distance between the personalization subset and test set reduces (Figure~\ref{fig:sfsDisteffect}). When we evaluated the NonRelapse-NonRelapse distance and NonRelapse-Relapse distance, i.e., the intra-class and inter-class distances, the difference in these distances was higher when personalization was used to subsample the training set (Figure~\ref{fig:sensordist_nonrel_rel_distribution}). This likely led to improved training as better differentiation between the classes was available for learning. Without personalization, more behavioral overlapped across classes could be present, for example, when one patient's normal behavior was more relapse-like for another patient. The effect of personalization was also seen with better differentiation between the relapse and non-relapse data points in the learned internal representation (embeddings) obtained from the trained \networkname (Figure~\ref{fig:tsne_personal_nonPersonal}).

Different mobile sensing modalities capture various aspects of one's behavioral patterns. To understand which modality is most helpful for relapse prediction, we trained \networkname with a single modality. The conversation and the volume modality provided the best relapse prediction performance (Table~\ref{tab:pred_single_modality}). Language/speech distortions and sociability impairments are implicated in schizophrenia \cite{delisi2001speech,kuperberg2010language, goldberg2001shyness, stanghellini2002dis}. The conversation and volume feature likely represents behavioral changes related to speech and sociability (based on how often conversations are happening), leading to better prediction performance. Other modalities such as the accelerometer resulted in a lower F2 score. In our relapse prediction model, we used the hourly averaged accelerometer magnitude as input to the model. The accelerometer signal is linked to behavior through different intermediaries such as activities that might not be represented by hourly averages. Therefore, it might be helpful to further process the accelerometer signal into representations such as total sedentary time, activities, etc. Such intermediate data transformations could be helpful for other modalities too and need to be explored in future work.

The supervised deep learning model, \networkname, performed better than an encoder-decoder architecture-based anomaly detection model for relapse prediction, each trained using hourly mobile sensing data. The anomaly detection model based on the encoder-decoder architecture uses unsupervised learning for embeddings generation from the mobile sensing data. A supervised step in identifying the best threshold that separates embedding from relapse phases and those from non-relapse phases is also used. However, the supervision is represented in a single scalar threshold. The \networkname, on the other hand, is highly parameterized through the weights of the LSTM and fully connected layers. Each of these weights is learned in a supervised way. The lower F2 score in the anomaly detection model for relapse prediction could be because the relapse behaviors are highly diverse and this diversity is not represented well with a single distance threshold parameter. 

\subsection{Limitations}

Though \networkname performed better than the existing deep learning model for relapse prediction, further improvements are highly likely, given the low F2 score in the full test set. One of the limitations of our work, commonly shared in many health applications, is the limited size of the dataset. Both a larger number of patients and a longer monitoring period could be helpful for the development of robust prediction models. A dataset with more patient, for example, will likely allow better model training since the personalization subset can be bigger. In this work, we used static demographic and mental health scores for personalization, thus accounting for some inter-individual differences in behavioral patterns. There might also be larger intra-individual differences. Therefore, a data-driven personalization where the personalization set is adapted to each data point/prediction points could be helpful. However, such personalization would also have higher computational costs and need considerations. Finally, we explored only simple late fusion approaches to combine the deep learning-based \networkname and the non-deep learning model-based \handcraftedmodel. Another predictor that learns a better combination of the predictions from the individual models could be helpful and needs to be explored in future work.

\section{Conclusion}

Relapse prediction using mobile sensing data could help provide timely interventions to schizophrenia patients. Previous work on deep learning models for relapse prediction using mobile sensing data evaluated unsupervised learning of features from raw sensor data using an encoder-decoder architecture. In this work, we investigated a fully supervised deep learning model with personalization and found it to outperform the encoder-decoder model. The deep learning model also complemented the non-deep learning model for relapse prediction. Thus, our work shows that a supervised deep learning model could be relevant for relapse prediction tasks. Personalization was found to be crucial for relapse prediction as this likely addresses inter-individual behavioral differences. The similarity of the patients defined according to their social functional scale (SFS) score provided the best personalization for a relapse prediction model. Our personalization approach only acknowledges the inter-individual behavioral differences and opens avenues for investigating personalization approaches to address intra-individual differences too in future work.


\ifCLASSOPTIONcaptionsoff
  \newpage
\fi



\bibliographystyle{IEEEtran}
\bibliography{IEEEabrv,schizophrenia_relapse}
%

%




\end{document}